%% file: neurips_2025.tex
\documentclass{article}

\usepackage{subcaption}
\usepackage{multirow}
 \usepackage{amsmath} 
 \usepackage{tabularx}
\usepackage{makecell}

\usepackage[final]{neurips_2025}

\usepackage{booktabs}
\usepackage{multirow}
\usepackage{colortbl}
\usepackage{xcolor}
\usepackage{amsmath}
 \usepackage{graphicx}
\usepackage[utf8]{inputenc} 
\usepackage[T1]{fontenc}    
\usepackage{hyperref}       
\usepackage{url}            
\usepackage{booktabs}       
\usepackage{amsfonts}       
\usepackage{nicefrac}       
\usepackage{microtype}      
\usepackage{xcolor}         

\title{Universal Camouflage Attack on Vision-Language Models for Autonomous Driving}

\author{%
  Dehong Kong\textsuperscript{1}\thanks{Equal contribution.} \quad
  Sifan Yu\textsuperscript{1}\footnotemark[1] \quad
  Siyuan Liang\textsuperscript{2} \quad
  Jiawei Liang\textsuperscript{1} \\
  \textbf{Jianhou Gan}\textsuperscript{3} \quad
  \textbf{Aishan Liu}\textsuperscript{4} \quad
  \textbf{Wenqi Ren}\textsuperscript{1}\thanks{Corresponding author.} \\
  \textsuperscript{1} School of Cyber Science and Technology, SUN YAT-SEN UNIVERSITY \\
  \textsuperscript{2} School of Computing, National University of Singapore \\
  \textsuperscript{3} Key Laboratory of Education Informatization for Nationalities, Yunnan Normal University \\
  \textsuperscript{4} SCSE, Beihang University \\
}

\begin{document}

\maketitle

\input{sec/0_abstract}    
\input{sec/1_intro}

\input{sec/2_related}

\input{sec/3_method}

\input{sec/4_experiment}
\input{sec/5_conclusion}

\clearpage
{

\small
\bibliography{main}
\bibliographystyle{plain}
}

\newpage

\end{document}

%% file: sec/0_abstract.tex
\begin{abstract}

Visual language modeling for automated driving (VLM-AD) is emerging as a promising research direction with substantial improvements in multimodal reasoning capabilities. Despite its advanced reasoning abilities, VLM-AD remains vulnerable to serious security threats from adversarial attacks, which involve misleading model decisions through carefully crafted perturbations. Existing attacks have obvious challenges: 1) Physical adversarial attacks primarily target vision modules. They are difficult to directly transfer to VLM-AD systems because they typically attack low-level perceptual components. 2) Adversarial attacks against VLM-AD have largely concentrated on the digital level. They suffer from significant limitations in real-world deployment, including their lack of physical realizability and sensitivity to environmental variability.
To address these challenges, we propose the first Universal Camouflage Attack (UCA) framework for VLM-AD. Unlike previous methods that focus on optimizing the logit layer, UCA operates in the feature space to generate physically realizable camouflage textures that exhibit strong generalization across different user commands and model architectures. Motivated by the observed vulnerability of encoder and projection layers in VLM-AD, UCA introduces a feature divergence loss (FDL) that maximizes the representational discrepancy between clean and adversarial images. In addition, UCA incorporates a multi-scale learning strategy and adjusts the sampling ratio to enhance its adaptability to changes in scale and viewpoint diversity in real-world scenarios, thereby improving training stability. Extensive experiments demonstrate that UCA can induce incorrect driving commands across various VLM-AD models and driving scenarios, significantly surpassing existing state-of-the-art attack methods (improving 30\% in 3-P metrics). Furthermore, UCA exhibits strong attack robustness under diverse viewpoints and dynamic conditions, indicating high potential for practical deployment.
\end{abstract}

%% file: sec/1_intro.tex
\section{Introduction}
\label{sec:intro}

\begin{figure}[htbp]
    \centering

    \begin{minipage}[t]{0.51\linewidth}
    \vspace{0pt}
        \begin{minipage}[t]{\linewidth}
        \scriptsize
            \centering
\begin{tabularx}{\linewidth}{c|c|c|c|c}
        \toprule
 \textbf{\makecell{Attack\\Type}} & \textbf{Physical} & \textbf{\makecell{Agnostic to\\Text}} & \textbf{VLM-AD} & \textbf{\makecell{Attack\\Level}} \\
        \hline
        Digital & $\times$ & $\times$ & $\checkmark$ & - \\
        \hline
        Patch & $\checkmark$ & - & $\times$ & logit \\
        \hline
        \makecell{General\\Camouflage}  & $\checkmark$ & - & $\times$ & logit \\
        \hline
        \rowcolor{gray!20}
        UCA & $\checkmark$ & $\checkmark$ & $\checkmark$ & feature \\
        \bottomrule
    \end{tabularx}
            
            \caption{Characteristic comparison with other attack methods.}
            \label{fig:1}
        \end{minipage}
\vspace{0.1cm} 

        \begin{minipage}[t]{\linewidth}
            \centering
            \includegraphics[width=1\linewidth]{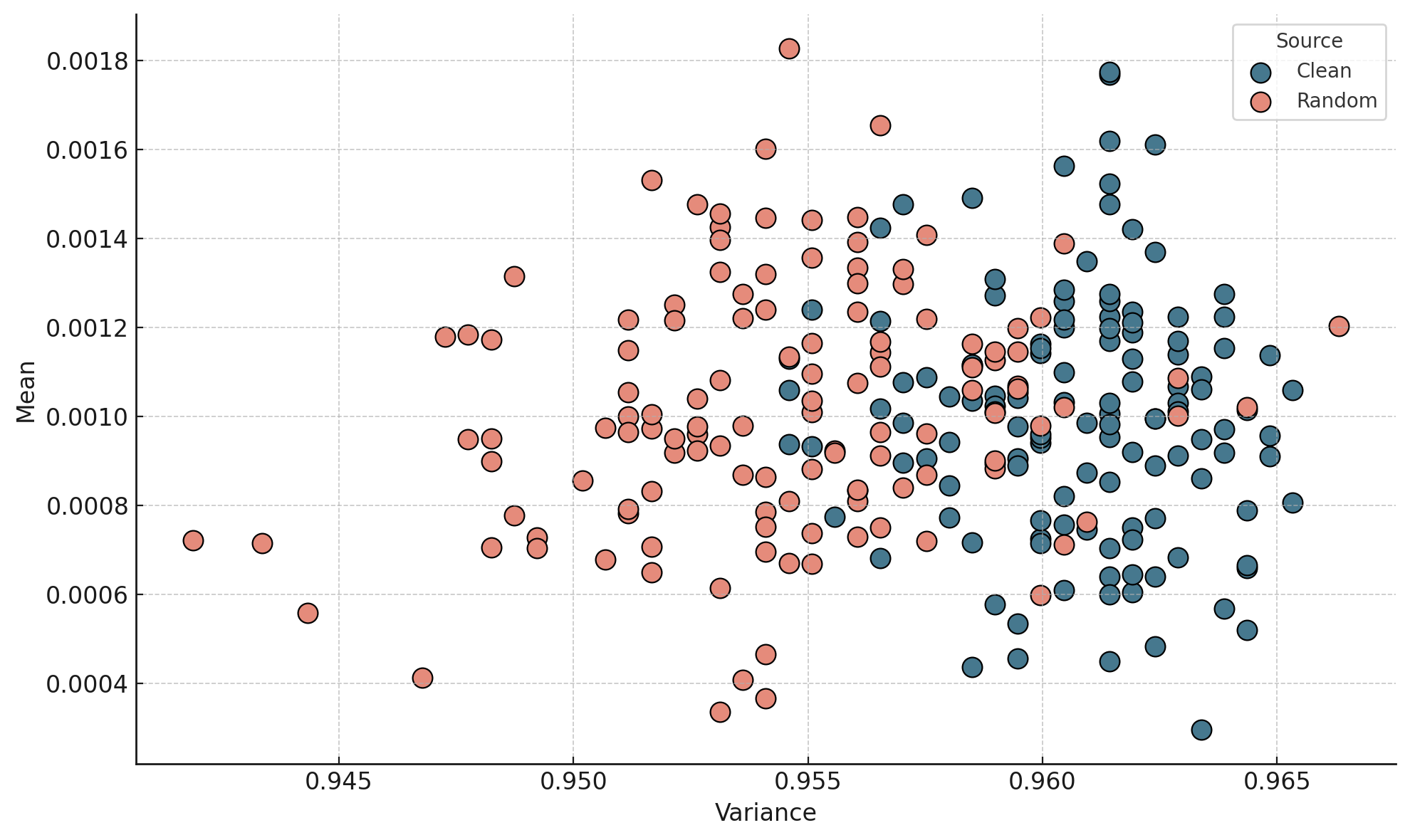}
            \caption{Feature distribution of projector between clean and random image. }
            \label{fig:2}
        \end{minipage}
    \end{minipage}
\hfill 
    \begin{minipage}[t]{0.45\linewidth}
    \vspace{0pt}
        \centering
        \includegraphics[width=\linewidth]{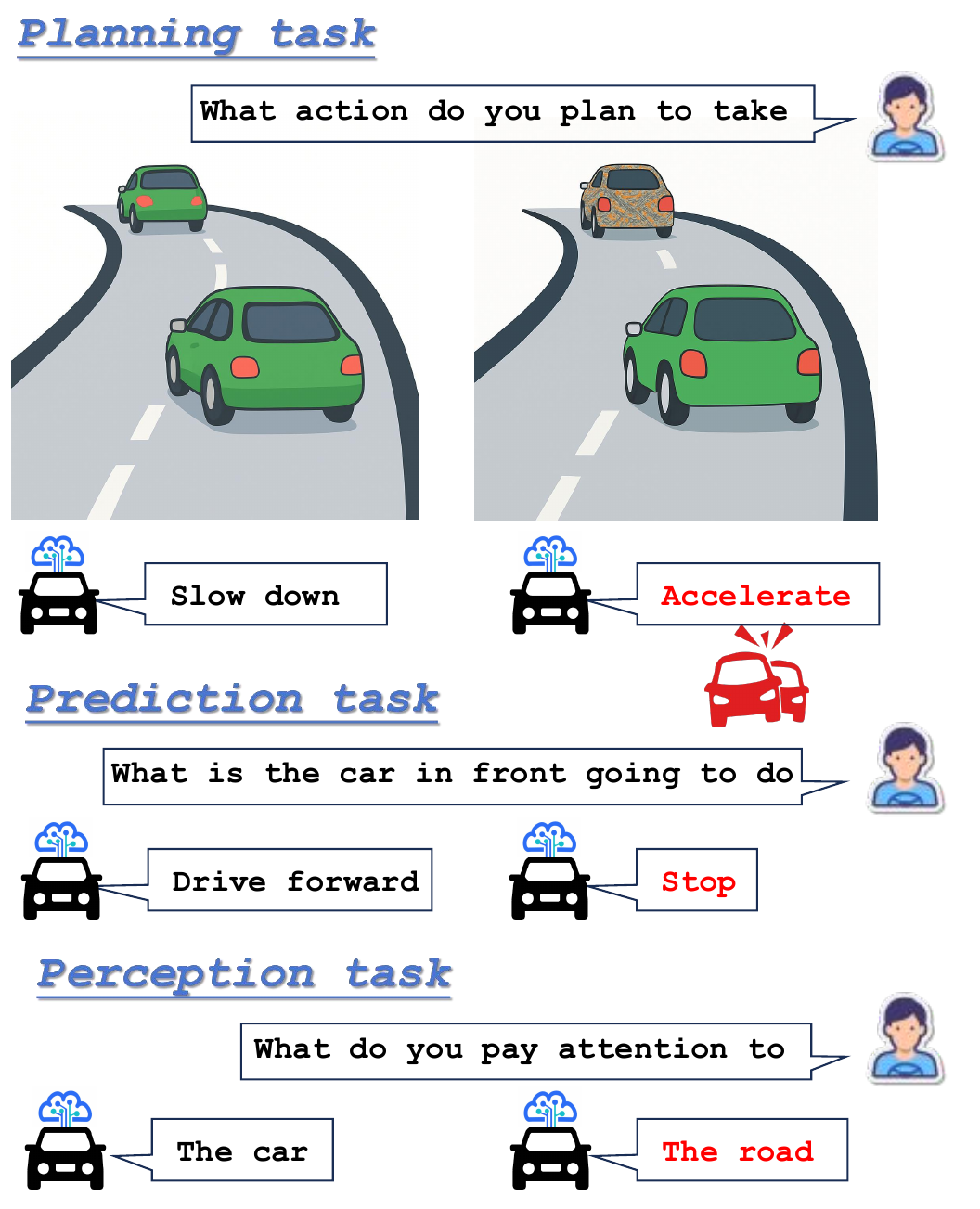}
        \vspace{-0.5cm}
        \caption{Our method are universal to various scenarios and mislead VLM-AD to generate wrong instructs.}
        \label{fig:3}
    \end{minipage}

\end{figure}

The application of Vision-Language Models (VLMs) to autonomous driving (i.e., VLM-AD) is emerging as a powerful new paradigm capable of fusing visual perception and language comprehension for multimodal reasoning and decision making.
Such models enable self-driving vehicles to understand driving commands, generate interpretable operational decisions, and interact with humans in natural language. 
However, the multimodal architecture of VLM-AD also introduces new safety risks~\cite{xiao2025genderbias, wang2025manipulating, liang2023badclip, liu2023x, liang2025revisiting, liang2025vl, ho2024novo}. 
Recent studies have shown that VLM-AD systems are particularly sensitive to adversarial attacks, which can cause serious safety hazards in real-world environments, such as predicting and planing through elaborate perturbation misdirection models \cite{zhang2024visual,wang2025black}.

Although research has begun to focus on adversarial robustness and some methods have been proposed to attack autonomous driving \cite{wang2023does,zhou2024stealthy,ma2024slowtrack,wang2024attack,wang2025unified,wang2025physically}, there are still significant limitations to the applicability of current attack methods in VLM-AD scenarios. Figure \ref{fig:1} shows the main difference among the attack methods of VLM-AD.
On the one hand, physical adversarial attacks~\cite{eykholt2018robust,10497142,zheng2024physical,wang2022fca,Wang_2021_das,suryanto2023active,zhou2024rauca} typically rely on optimizations to the logit layer of the model, aiming to manipulate the probability of category prediction, object location, or specific tokens. Such approaches are more effective in traditional vision tasks, such as object detection, due to their relatively simple output structure. However, in VLM-AD scenarios, the output of the model is often a natural language instruction consisting of multiple tokens with a high degree of semantic complexity and free generation. As a result, the attack of such low-level perturbations on high-level semantic decisions is extremely limited.
On the other hand, most of the existing attacks against VLM-AD focus on the digital level and mislead mainly by manipulating textual inputs. But such methods lack physical realizability and are less robust in the face of real-world changes in lighting, scale, and perspective. 
Thus, these limitations indicate that there is still a lack of a physically realizable and semantically effective adversarial attack for VLM-AD at this stage.

In this paper, we propose the first Universal Physical Camouflage Attack framework, termed Universal Camouflage Attack (UCA), for VLM-AD systems. 
Unlike digital attacks that rely on the logit layer to manipulate the output vocabulary, UCA launches the attack directly in the feature space, interfering with the visual representations of the encoder and projector layers in VLM-AD, thus disrupting the model's multimodal semantic modeling process and realizing a universal physical attack across tasks (see Figure \ref{fig:3}). 
In addition, compared with the localized patch-based attack, UCA optimizes the camouflage texture of the entire vehicle surface, which has stronger viewpoint adaptability and physical deployability.

First, we find that the encoder and projector layers in VLM are highly sensitive to visual texture variations. 
Experiments show that the mean and variance of the output features of the projector layer are significantly altered after a random texture is applied to the vehicle surface (see Figure \ref{fig:2}). 
Considering that mainstream VLM models generally adopt a similar encoding architecture, this observation suggests shifting from the logit layer to a more generalized feature representation layer to perform attacks and generate adversarial textures that can be transferred across different architectures. Second, we design two key optimization strategies to improve the stability and practicality of the attack against inevitable viewpoint variations and scale shifts in physical deployment. 
For viewpoint differences, we adopt a perspective reweighted sampling mechanism to enhance the contribution of samples from attack-sensitive viewpoints during training, thereby improving the overall attack efficacy. 
In addition, we introduce a multi-scale training strategy, inspired by the scale modeling techniques in small object detection, to ensure the model maintains effective attack responses at long distances or under target size reduction. 
The synergistic effect of these two strategies effectively enhances the robustness and attack success rate of UCA in complex real-world scenarios.

Experimental results show that our generated universal camouflage texture is able to simultaneously and effectively attack the three mission-critical modules of perception, prediction, and planning. Compared with existing state-of-the-art approaches, the overall attack success rate of UCA is improved by more than 30\%. The main contributions are listed as follows:
\begin{itemize}
    \item We are the first to extend adversarial camouflage attacks to the physical domain of VLM-AD, enabling cross-task universal attacks beyond prior digital-level methods.
    \item We propose a novel feature-space attack that targets the encoder and projector layers of VLM-AD, and further enhance physical robustness through a reweighted sampling strategy and multi-scale training to address real-world viewpoint and scale variations.
    \item Extensive experiments on perception, prediction, and planning tasks demonstrate the superior effectiveness and generalizability of our method across diverse scenarios.
\end{itemize}

%% file: sec/2_related.tex
\section{Related Work}
\label{sec:related}

\setlength{\parskip}{1em}

\subsection{VLMs in Autonomous Driving}
Recent advances in large language models (VLMs) have expanded their applications in autonomous driving. 
DiLu \cite{wen2023dilu} and GPT-Driver\cite{mao2023gpt} explored using GPT-3.5 and GPT-4 as planning modules for driving decisions.
Later work \cite{xu2024drivegpt4,huang2024makinglargelanguagemodels} proposed end-to-end LMM-based frameworks that directly generate control commands or driving trajectories.
In contrast to language-driven approaches, models like \cite{shao2024lmdrive,wang2023drivemlm} employ decoders to infer control signals from latent representations.
To further improve perception and reasoning capabilities, various architectural innovations\cite{nie2025reason2drive,ding2024holistic,zhou2024embodied} have been introduced.
Despite their promising results, most existing methods are constrained to specific driving scenarios or tailored tasks, such as particular datasets or data formats.
CODA-LM~\cite{chen2024automated} introduced an automated benchmark for long-tail driving scenarios, using text-based VLMs for evaluation and demonstrating enhanced decision analysis via structured prompts. OmniDrive~\cite{wang2024omnidrive} proposed a sparse query-based architecture for 3D scene modeling, integrating dynamic-static object representation with memory-enhanced positional encoding. 
We follow their evaluation metrics to compare attack performance.
Dolphins~\cite{ma2024dolphins} first enhance reasoning capabilities through an innovative Grounded Chain of Thought (GCoT) process. Then the authors tailored Dolphins to the driving domain by constructing driving-specific instruction data and conducting instruction tuning. Through the utilization of the BDD-X dataset, they designed and consolidated four distinct AV tasks into Dolphins to foster a holistic understanding of intricate driving scenarios. We choose Dolphins as our main victim model.
\subsection{Physical Adversarial Attacks}
Physical adversarial attacks manipulate object characteristics to deceive vision systems, categorized into patch-based and camouflage-based approaches. Patch-based methods apply localized adversarial patterns to object surfaces or backgrounds. Those methods are mainly designed to attack object detectors ~\cite{kong2024patch,jing2024pad,wang2025unified,long2024papmot,wei2025real,guesmi2024dap,wei2024revisiting,li2024physical,ran2023cross}. DGA ~\cite{10497142} propose a new direction-guided attack to deceive real-world aerial detectors.
However, their planar constraints limit robustness under multi-view or long-distance conditions. Camouflage-based methods enhance stealth by optimizing 3D textures or shapes ~\cite{xia2025alphadog,ziqing2024camouflage,wang2025highly,huang2024uba,zhang2025phycamo,peng2025physical,zhu2025camoenv,hu2021cca,wang2023sc}. FCA~\cite{wang2022fca} introduced Full-coverage Camouflage Attack, which maps adversarial textures onto entire vehicle surfaces using neural rendering and environmental transformations to address multi-view failures. 
the Dual Attention Suppression (DAS) attack\cite{Wang_2021_das} reduces visibility to both models and human observers.  
Some works \cite{suryanto2022dta,suryanto2023active} employe a neural renderer to simulate realistic effects like shadows and improve camouflage aesthetics with enhanced texture mapping and background color integration. Furthermore, RAUCA\cite{zhou2024rauca} utilizes
advanced rendering to account for environmental factors such as diverse weather conditions. 
Although these physical attack methods show some effectiveness in specific perceptual tasks, they generally rely on fixed viewpoints, specific tasks, or model outputs (e.g., detection boxes or depth) and are difficult to be extended to complex systems requiring multimodal semantic reasoning such as VLM-AD. Our UCA is the first camouflage attack against VLM-AD.

\subsection{ Adversarial Attacks on VLM-AD}
Adversarial attacks on VLMs for autonomous driving systems have attracted significant attention, focusing on dynamic scene adaptability, multimodal vulnerabilities, and robustness in safety-critical scenarios. Zhang et al. ~\cite{zhang2024visual} developed ADvLM, employing semantic-invariant induction to create instruction libraries and scene-correlation optimization for temporal perturbations, enhancing attack robustness across dynamic perspectives. For black-box scenarios, Wang et al. ~\cite{wang2025black} proposed the Cascaded Adversarial Disturbance (CAD) framework, inducing cross-reasoning-chain errors via decision-chain disruption and risk-scenario induction in dynamic environments. 
Although existing adversarial attacks against VLM-AD have made some progress in dynamic scenario modeling and multimodal vulnerability, they are still largely confined to the digital level, limiting their threat realism.

%% file: sec/3_method.tex
\section{Method}
\label{sec:method}
\begin{figure}[tbp]
    \centering
    \includegraphics[width=\textwidth]{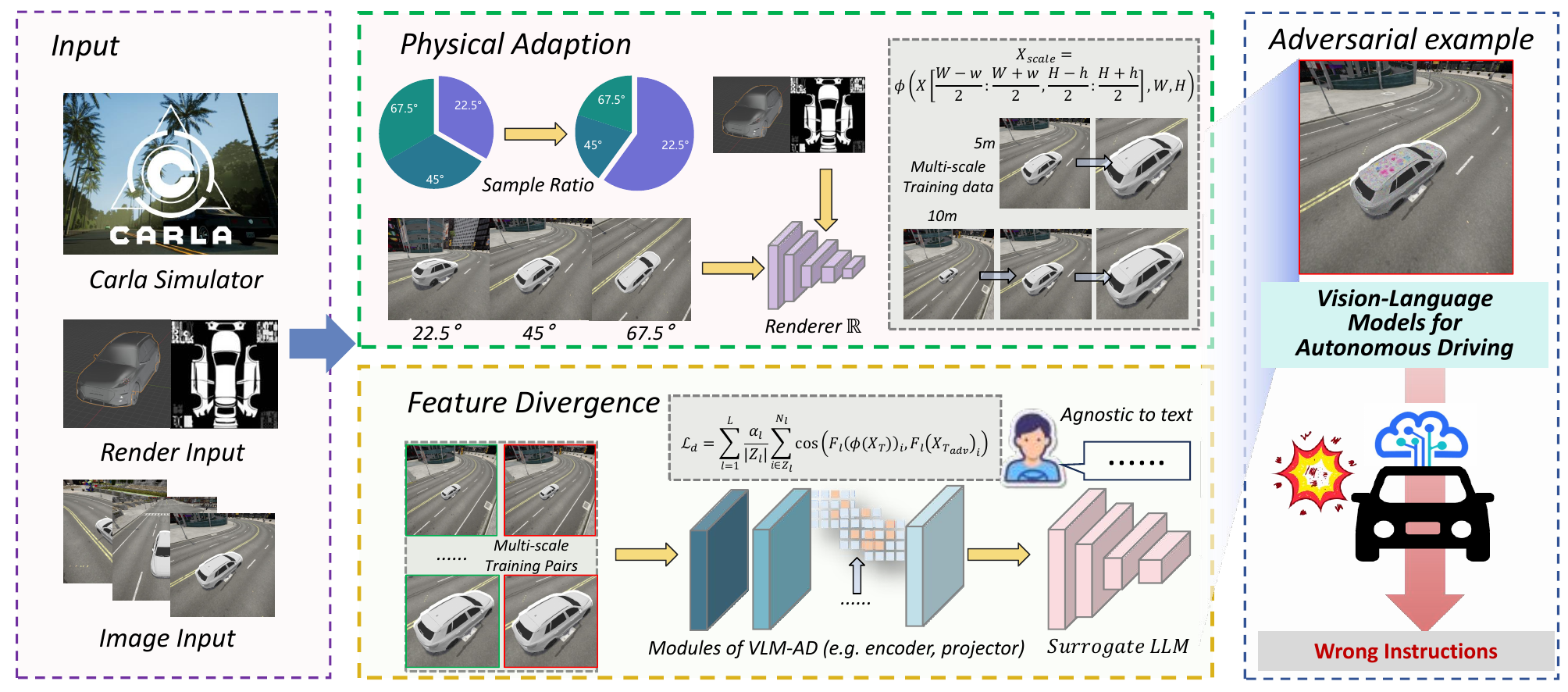} 
    \caption{Attack Framework UCA. Our approach introduces Feature Divergence, which targets the feature distribution across multi-layers of VLM-AD. UCA proposes a new sampling strategy and leverages multi-scale prior to adapt physical environment.}
    \label{fig:pipe}
\end{figure}

Figure \ref{fig:pipe} shows the framework to attack VLM-AD. The attack scheme is to generate the adversarial camouflage texture utilizing the neural renderer to paint on the surface of the 3D vehicle model. Based on the analysis of the vulnerability of VLM, we manipulate the intermediate features of modules of VLM (e.g. vision models, projectors) to disturb the output of models.

\subsection{Problem Formulation}
Given a training dataset ($\textbf{X}$,$\theta_c$) where $\textbf{X}$ 
and $\theta_c$ are the sampled images and the corresponding camera parameters respectively, a 3D car model with a mesh $\textbf{M}$ and a texture \textbf{T}, 2D car image $\textbf{O}$ can be generated by a renderer $\mathcal{R}$:
\begin{equation}
    \textbf{X}_{\mathrm{T}} = \mathcal{R}(\textbf{M},\textbf{T};\theta_c).
\end{equation}
To realize the adversarial camouflage attack, we replace the original texture $\textbf{T}$ with adversarial texture $\textbf{T}_{\mathrm{adv}}$ and obtain the adversarial image $\textbf{X}_{\mathrm{T}_{\mathrm{adv}}}$ with transformation function $\phi$. We aim to input ($\textbf{X}_{\mathrm{adv}}$,t) to attack $\mathcal{F}$ to output the wrong text or reduce its performance, where $\mathcal{F}$ is VLM-AD and t is the benign text input.

We treat the manipulation as an optimization problem, and the function is expressed as follows
\begin{equation}
    \hat{\textbf{T}}_{\mathrm{adv}} = \mathop{\arg\max}\limits_{\mathrm{T_{adv}}} \mathcal{J}(\mathcal{F}(\phi(\textbf{X}_{\mathrm{T}}),t),\mathcal{F}(\phi(\textbf{X}_{\mathrm{T}_{\mathrm{adv}}}),t)),
\end{equation}
where $\hat{\textbf{T}}_{\mathrm{adv}}$ is the trained adversarial texture,$t$ is the textual input from users, and $\mathcal{J}(\cdot,\cdot)$ is the loss function.

We generate the adversarial camouflage texture by utilizing a differentiable neural renderer. It enables the direct application of customized textures onto 3D car models. This is the first attempt in the field of autonomous driving adversarial attacks.

\subsection{Targeted Feature Divergence for Universal VLM Attacks}

To enable universal adversarial camouflage across diverse downstream tasks, we propose a task-agnostic \textbf{feature divergence minimization} strategy that disrupts multi-layer visual representations. Specifically, we aim to perturb the texture map \( \mathbf{T} \) of a 3D mesh such that the rendered image \(\phi(\mathbf{X}_{\mathrm{T_{adv}}})\), after transformation $\phi$ (see \ref{pyhsical}), exhibits maximal deviation from its benign counterpart \(\phi(\mathbf{X}_{\mathrm{T}})\) in a task-agnostic feature space across multiple layers.

To identify the most vulnerable visual patterns under adversarial perturbation, we define a set of \textbf{key features} at each feature layer \( l \in \{1, \dots, L\} \), denoted by \( Z_l \), which are selected based on their cosine similarity difference under transformation \( \phi \):
\begin{equation}
    Z_l = \left\{ i \mid \cos\left( F_l(\phi(\mathbf{X}_{\mathrm{T}}))_i, F_l(\phi(\mathbf{X}_{\mathrm{T_{adv}}}))_i \right) \leq \delta \right\},
\end{equation}
where \( \delta \) is a threshold for selecting significantly deviated features, \( F_l(\cdot) \) denotes the feature representation extracted at layer \( l \), and \( \mathrm{cos}(\cdot) \) denotes cosine similarity. 

We then enforce divergence between these important features by minimizing the aggregated cosine similarity across layers:
\begin{equation}
    \mathcal{L}_{d} = \sum_{l=1}^{L} \frac{\alpha_l}{|Z_l|} \sum_{i \in Z_l} \cos\left( F_l(\phi(\mathbf{X}_{\mathrm{T}}))_i, F_l(\phi(\mathbf{X}_{\mathrm{T_{adv}}}))_i \right),
\end{equation}
where \( \alpha_l \) is a weighting factor for each layer and \( |Z_l| \) is the number of selected features at layer \( l \).
 
 Combined with a differentiable renderer and transformation module \( \phi \), our approach produces adversarial textures \( \mathbf{T}_{adv} \) that are robust and transferable under real-world augmentations.

 For any user textual input \( t \in \mathcal{T} \), the condition for a successful attack is:
\[
\forall t \in \mathcal{T}, \quad \mathcal{F}(\phi(\mathbf{X}_{\mathrm{T}_{\mathrm{adv}}}), t) \neq \mathcal{F}(\phi(\mathbf{X}_{\mathrm{T}}), t).
\]
In other words, the adversarial texture \( \mathbf{T}_{\mathrm{adv}} \) induces universal erroneous predictions by the model, regardless of the input text.

\subsection{Adaptive View-Scale Sampling for Physical Robustness}
\label{pyhsical}
In real-world physical environments, the angle and height of the camera or sensor often vary, introducing additional complexities for feature extraction.
In the previos method, the sampling strategy followed a balanced distribution with an equal ratio of \( 1:1:1 \) for the angles \( 22.5^\circ \), \( 45^\circ \), and \( 67.5^\circ \). However, based on empirical observations, it was found that this approach led to suboptimal results, particularly in the case of the \( 22.5^\circ \) angle, where adversarial attacks often failed. This suggests that the sampling at some angle did not capture the important features effectively, making the model vulnerable to adversarial perturbations and different region of texture have different training difficulties.

To address this issue, we proposed to shifts the sampling ratio from \( 1:1:1 \) to \( 3:1:1 \), giving more weight to the \( 22.5^\circ \) angle. This adjustment enhances the sampling density around the critical angle, improving the model's robustness against adversarial attacks and leading to better overall performance. This adjustment ensures that the sampling process now prioritizes spercific angle more heavily, addressing its previous vulnerabilities in adversarial scenarios.

Inspired by small object detection techniques, we propose multi-scale strategy to address the challenge from distance.  Experimental observations showed that an image at 5m distance yielded better results compared to a 10m distance, where the texture features become less prominent, leading to difficulties in accurate detection and adversarial robustness.
 This strategy adapts feature representations of \( X \) at different scales, ensuring better detection and robust adversarial feature extraction.

Mathematically, the process is described as follows:
\begin{equation}
    X_{\text{scale}} = \phi(X_{\mathrm{crop}}, W, H),
\end{equation}
\begin{equation}
X_{\text{crop}} = X\left[\frac{W - w}{2}: \frac{W + w}{2}, \frac{H - h}{2}: \frac{H + h}{2}\right],
\end{equation}
where \( X \) represents the original image, \( W \) and \( H \) are the original dimensions of \( X \), \( \phi \) is the rescaling function that resizes the image,
 \( w \) and \( h \) represent the desired width and height of the cropped region,\( \frac{W - w}{2} \) and \( \frac{H - h}{2} \) define the starting points for cropping from the center. By cropping and resizing, images of different scales can be obtained, thereby improving the training efficiency of images at longer distances.

To ensure the naturalness of the generated adversarial camouflage, we utilize the smooth loss to reduce the inconsistency among adjacent pixels. For a rendered car image painted with adversarial camouflage $\textbf{X}_{adv}$, the calculation of smooth loss can be written as
\begin{equation}
    \mathcal{L}_{\text{smooth}} = \sum_{i,j} \left( (x_{i,j} - x_{i+1,j})^2 + (x_{i,j} - x_{i,j+1})^2 \right),
\end{equation}
where $x_{i,j}$ is the pixel value of $\textbf{X}_{adv}$ at coordinate $(i,j)$.

\subsection{Framework of Universal Camouflage Attack}




Our framework optimizes an adversarial texture \(\mathbf{T}_{adv}\) to generate robust and physically camouflage textures that maintain its effectiveness across diverse viewpoints and tasks. Specifically, a differentiable renderer \(\mathcal{R}\) synthesizes 2D images from a 3D model \(\mathbf{M}\) textured by \(\mathbf{T}_{adv}\), conditioned on camera parameters \(\theta_c\). To simulate real-world variations, we apply cropping transformations \(\theta_{\mathrm{crop}} \sim \phi_{\mathrm{crop}}\) and scaling \(\theta_{\mathrm{scale}} \sim \phi_{\mathrm{scale}}\).

We then optimize the adversarial texture by minimizing the expected loss over the input data distribution \(\mathcal{D}\) and the sampled transformations. Our objective consists of two components: the feature divergence loss \(\mathcal{L}_d\), which maximizes the discrepancy between the model’s multi-layer feature representations of clean and adversarial images, promoting universal mislead; and the smoothness loss \(\mathcal{L}_{smooth}\), which encourages spatial consistency in the adversarial texture to maintain natural appearance. A balancing hyperparameter \(\lambda_s\) controls the trade-off between these terms.

Formally, the optimization is expressed as:

\begin{equation}
\min_{\mathbf{T}_{adv}} \;  \mathbb{E}_{\substack{\theta_{\text{crop}} \sim \mathcal{\phi}_{\text{crop}} \\ \theta_{\text{scale}} \sim \mathcal{\phi}_{\text{scale}}}} \left[
\mathcal{L}_{\text{d}}\left(
\mathcal{\phi}(\mathrm{\mathcal{R}(\textbf{M},\textbf{T}_{adv};\theta_c)}; \theta_{\text{view}}, \theta_{\text{scale}})
\right)
+ \lambda_s \mathcal{L}_{\text{smooth}}(\mathbf{T}_{\mathrm{adv}})
\right].
\end{equation}

This framework enables the generation of adversarial camouflage that is both universal—effective regardless of specific task inputs—and physically robust, ensuring practical applicability in real-world autonomous driving scenarios.

%% file: sec/4_experiment.tex
\section{Experiment}
\label{sec:experiment}
\subsection{Experiment Setup}
We select SoTA VLM-based
AD models Dolphins for attack. Patch attack DGA\cite{10497142}, some camouflage attack DAS\cite{Wang_2021_das}, FCA~\cite{wang2022fca}, RAUCA\cite{zhou2024rauca} and are baseline methods. 
We follow the common practice metrics in relevant works for comparison. CODA-LM~\cite{chen2024automated} uses text-only VLMs,
\textit{e.g.} GPT-4, as evaluators to score model responses. OmniDrive~\cite{wang2024omnidrive} employs rule-based language metrics to evaluate sentence similarity at the word level. We propose the 3-P metrics, including planning, prediction, and perception, to measure the success rate of attacks.

We follow DAS and FCA to utilize photo-realistic datasets to
perform the experiments. We select the simulator CARLA  for AD research. The CARLA simulator provides a variety of
high-fidelity digital scenarios. Specifically,we use different distance
 values (5m and 10m), three camera pitch angle values
 (\( 22.5^\circ \) \,\( 45^\circ \), and \( 67.5^\circ \)), and eight camera yaw angle values (south, north, east, west, southeast, southwest, northeast, northwest). We then collect 6000 simulation
 images with different setting combinations for training. Learning rate and max epoch is 0.1 and 5, respectively. 

\subsection{Attack Effectiveness}

\begin{table}[t]
\footnotesize
\caption{Evaluation results on NLP and LLM Judge metrics.}
\vspace{0.3cm}
\centering
\setlength{\tabcolsep}{4.7pt}
\begin{tabular}{l|cccc|cccc}
\toprule
\multirow{2}{*}{\textbf{Methods}} & \multicolumn{4}{c}{\textbf{NLP Metrics}} & \multicolumn{4}{c}{\textbf{LLM Judge}} \\
\cmidrule(lr){2-5} \cmidrule(lr){6-9}
& BLEU & METEOR & ROUGE & Average & General & Regional & Suggestion & Average \\
\midrule
clean         & 1.0000 & 1.0000 & 1.0000 & 1.0000 & 10.0 & 10.0 & 10.0 & 10.0 \\
Random        & 0.5769 & 0.5335 & 0.6819 & 0.5974 & 8.2  & 8.5  & 7.9  & 8.2 \\
DGA\cite{10497142}           & 0.5194 & 0.4761 & 0.5960 & 0.5305 & 8.0  & 8.3  & 8.1  & 8.1 \\
DAS\cite{Wang_2021_das}           & 0.5439 & 0.5063 & 0.6144 & 0.5548 & 7.9  & 8.0  & 7.8  & 7.9 \\
FCA\cite{wang2022fca}          & 0.5658 & 0.5231 & 0.6253 & 0.5714 & 8.4  & 8.6  & 8.3  & 8.4 \\
RAUCA\cite{zhou2024rauca}         & 0.4689 & 0.4269 & 0.5294 & 0.4751 & 7.1  & 7.2  & 7.0  & 7.1 \\
\rowcolor{gray!20}
Our           & 0.3946 & 0.3486 & 0.4722 & \textbf{0.4051} & 4.3 & 4.5 & 4.1 & \textbf{4.3} \\
\bottomrule
\end{tabular}
\label{tab:1}
\end{table}

\begin{table}[tbp]
    \centering
     \caption{Evaluation results on 3-P metrics.}
     \vspace{0.3cm}
    \setlength{\tabcolsep}{3.5pt}
    \label{tab:comparison_methods}
    \begin{tabular}{l|c c c c c c c c}
        \toprule
        \cmidrule(r){2-9}
        & Clean & Random & Digital & DAS\cite{Wang_2021_das} & FCA\cite{wang2022fca} & DGA\cite{10497142} & RAUCA\cite{zhou2024rauca} & Our \\
        \midrule
        PLAINING & 0\% & 0\% & 4\% & 36\% & 26\% & 28\% & 14\% & 78\% \\

        PREDICTION & 0\% & 0\% & 2\% & 16\% & 24\% & 20\% & 40\% & 56\% \\

        PERCEPTION & 0\% & 0\% & 8\% & 5\% & 4\% & 10\% & 18\% & 28\% \\

        Average & 0\% & 0\% & 7\% & 22\% & 18\% & 19\% & 24\% & \textbf{54}\% \\
        \bottomrule
    \end{tabular}
    \vspace{0.3cm}
   
    \label{tab:2}
\end{table}

\begin{figure}[t!]
    \centering
    \begin{subfigure}[t]{0.16\textwidth}
        \centering
        \includegraphics[width=\linewidth]{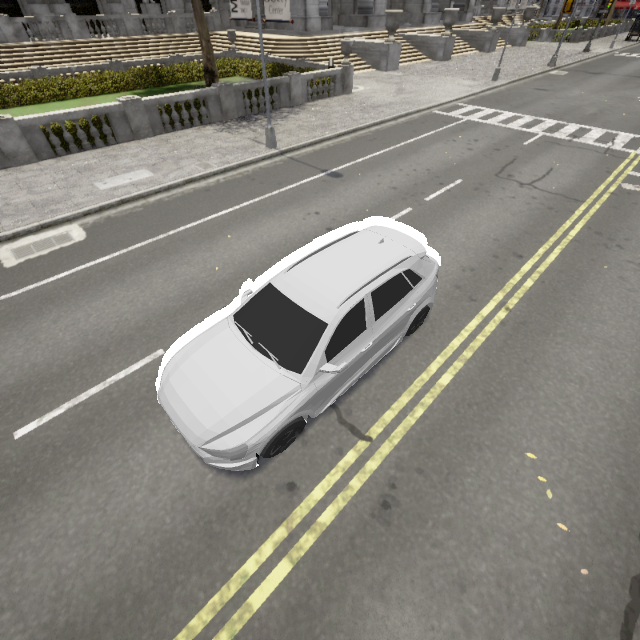}
        \caption{Clean}
 
    \end{subfigure}
    \begin{subfigure}[t]{0.16\textwidth}
        \centering
        \includegraphics[width=\linewidth]{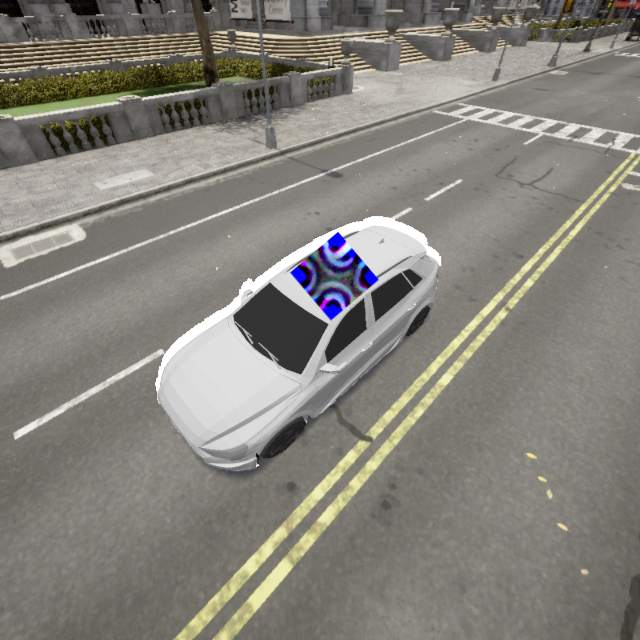}
        \caption{DGA}

    \end{subfigure}
    \begin{subfigure}[t]{0.16\textwidth}
        \centering
        \includegraphics[width=\linewidth]{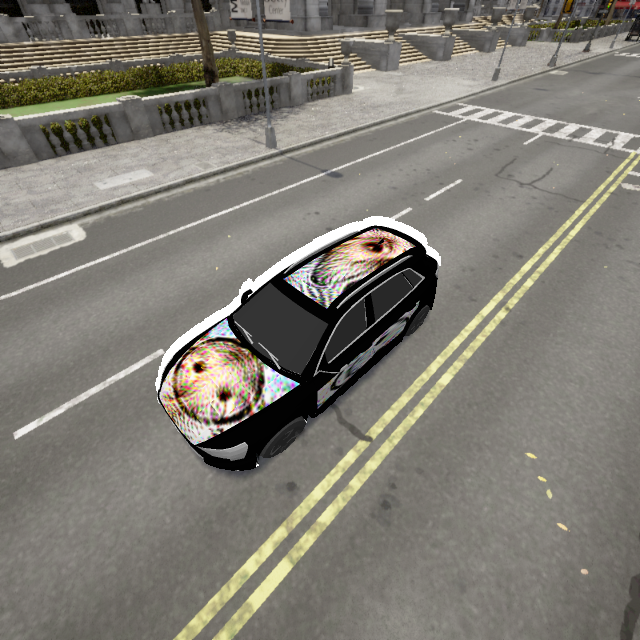}
        \caption{DAS}

    \end{subfigure}
    \begin{subfigure}[t]{0.16\textwidth}
        \centering
        \includegraphics[width=\linewidth]{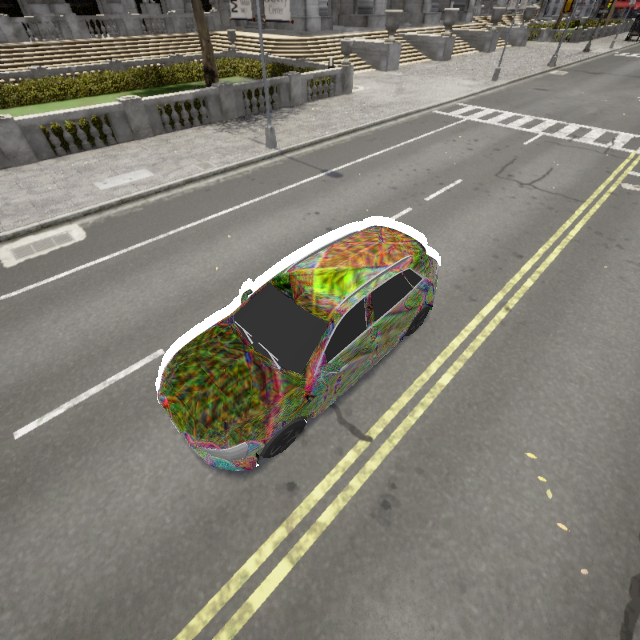}
        \caption{FCA}

    \end{subfigure}
    \begin{subfigure}[t]{0.16\textwidth}
        \centering
        \includegraphics[width=\linewidth]{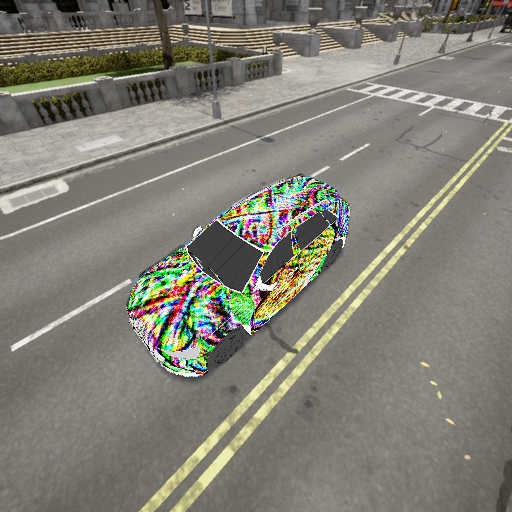}
        \caption{RAUCA}

    \end{subfigure}
    \begin{subfigure}[t]{0.16\textwidth}
        \centering
        \includegraphics[width=\linewidth]{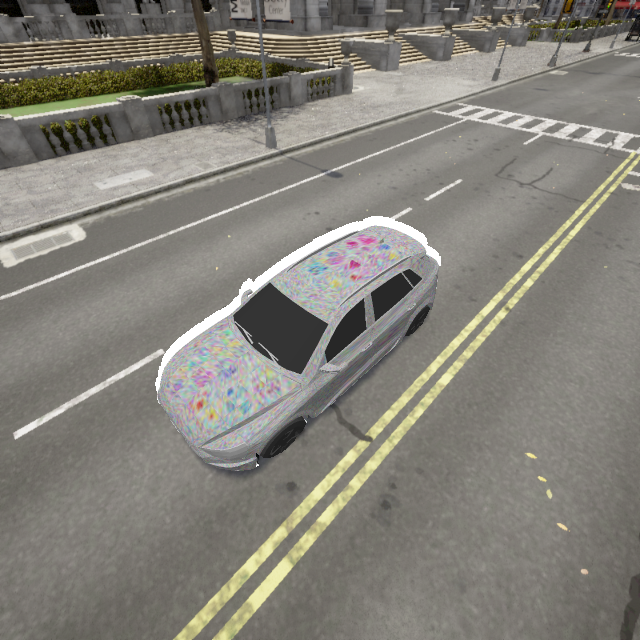}
        \caption{Our}

    \end{subfigure}
    \caption{Samples of different methods.}
    \vspace{-0.3cm}
    \label{fig:six_images}
\end{figure}

In this section, we evaluated the effectiveness of our proposed universal camouflage attack (UCA) against VLM-AD. The evaluation was conducted using two sets of metrics: NLP Metrics and LLM Judge scores, as detailed in Table \ref{tab:1}. Additionally, we assessed the attack success rate across three distinct scenarios, as summarized in Table \ref{tab:2}. Samples of different methods are shown in Figure \ref{fig:six_images}.

NLP Metrics directly assess the similarity between the generated text and the ground truth text. Lower values indicate better performance, as they signify higher divergence from the original content. The metrics used include BLEU, METEOR, and ROUGE, with their averages reported to provide an overall assessment. Our method achieves significantly lower scores compared to other approaches across all NLP metrics. Specifically, our method yields an average score of 0.4051 , which is notably lower than other methods. 

To evaluate the semantic quality and detectability of the generated adversarial text from an external perspective, we employed large language models, such as GPT-4 as independent judges. We prompted each LLM to score the output text along three dimensions: General, Regional, and Suggestion. More details can be found in the supplemental materials. Lower scores indicate that the text is less aligned with expected behavior, suggesting a more effective attack.
Our method also outperforms other methods. The average LLM Judge score for our method is 4.3, which is substantially lower than the scores obtained by competing methods. This suggests that our method successfully attack VLM-AD model, as evidenced by the low scores of three driving dimensions.

To further validate the robustness of our method, we constructed three driving scenarios (PLAINING, PREDICTION, and PERCEPTION) to measure the attack success rate. 
The overall average attack success rate across all three scenarios is 54\% , highlighting the effectiveness of our method in compromising the integrity of VLM-AD under diverse conditions. The high success rates observed in all three scenarios suggest that our method is universal across different driving scenarios and agnostic to text input.

\subsection{Attack Robustness}
\begin{figure}[t!]
    \centering
    \begin{subfigure}[t]{0.49\textwidth}
        \centering
        \includegraphics[width=\linewidth]{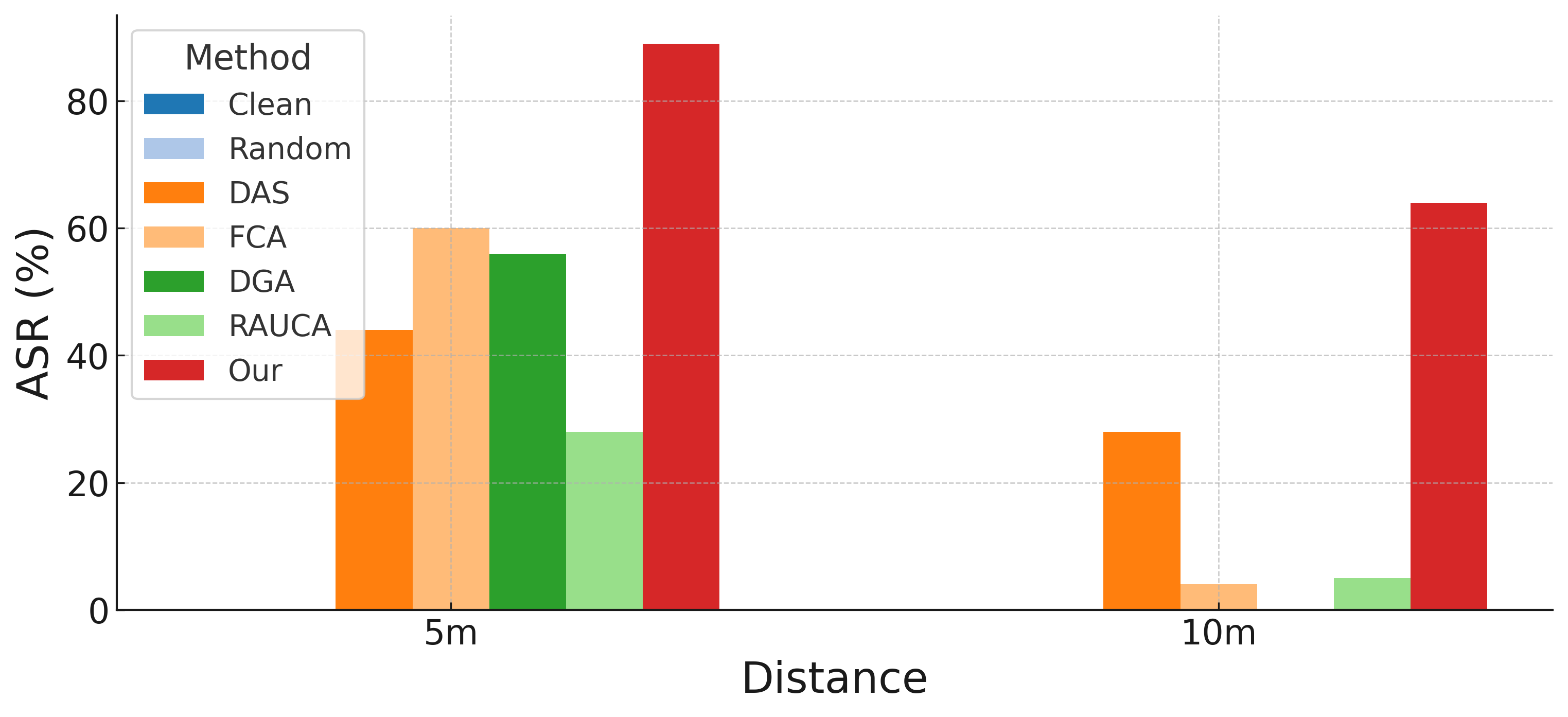}
        \caption{}
        \label{fig:sub3}
    \end{subfigure}
    \begin{subfigure}[t]{0.49\textwidth}
        \centering
        \includegraphics[width=\linewidth]{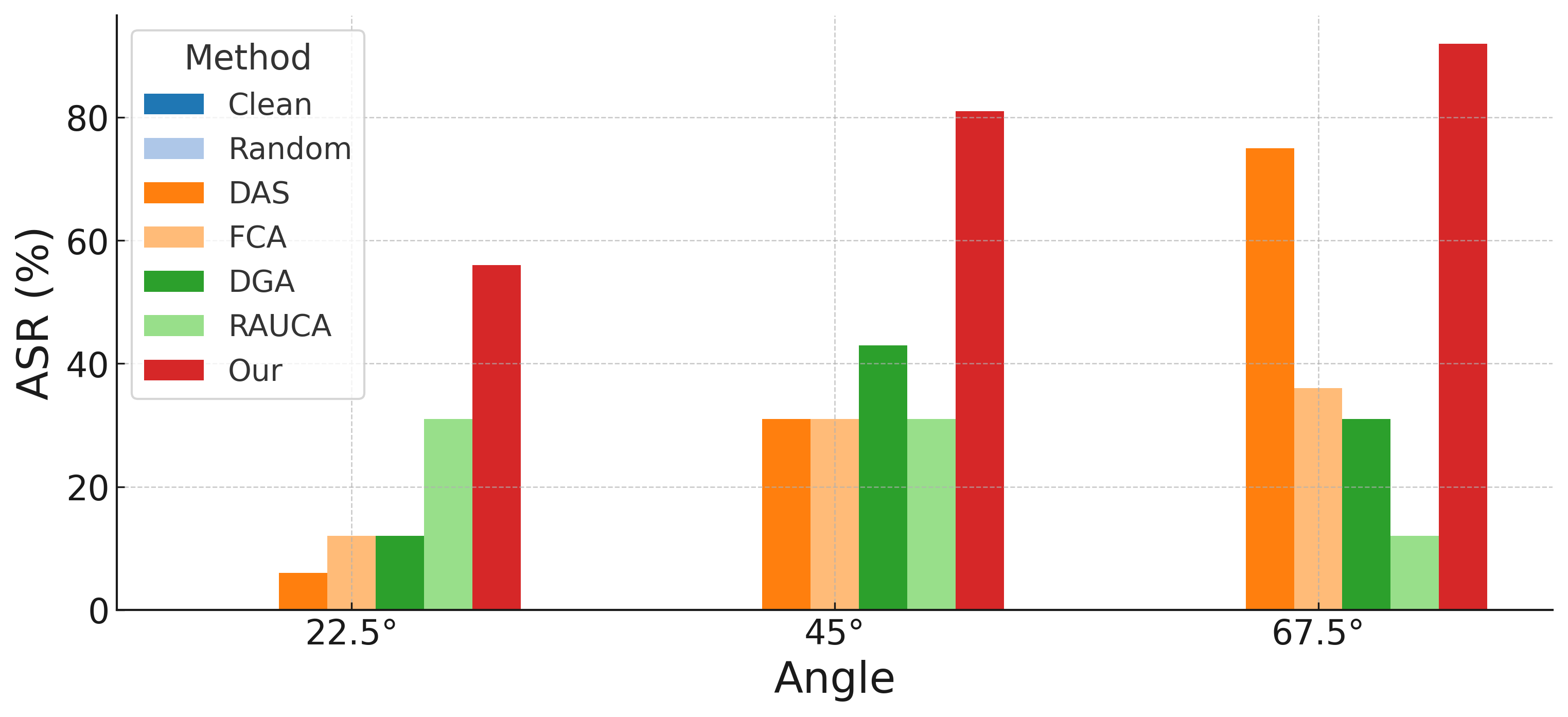}
        \caption{}
        \label{fig:sub4}
    \end{subfigure}

    \caption{Evaluation results in various distances and angles.}
    \vspace{-0.6cm}
\end{figure}

We evaluated the effectiveness of our attack under varying camera viewpoints, including different distances (5m and 10m) and pitch angles (22.5°, 45°, and 67.5°), to simulate realistic adversarial scenarios in autonomous driving perception systems. 

As shown in Figure \ref{fig:sub3}, our method significantly outperforms all baseline approaches at both distances. At a distance of 5m , our attack achieves an impressive success rate of 78\% , which is 30–40\% higher than most competing methods. Even at a farther distance of 10m , where adversarial perturbations become less effective due to reduced texture resolution, our approach still maintains a strong ASR of 56\% , demonstrating its robustness in long-range attack scenarios. From Figure \ref{fig:sub4}, it is evident that the attack performance degrades with smaller pitch angle. However, even under challenging conditions such as a 22.5° angle, our method still performances best. Our method explicitly considers variations in camera angle during training and incorporates multi-scale priors, allowing the attack to remain effective from multiple perspectives.

\subsection{Discussion with Digital Attack}
While digital attack can degrade the performance of VLM-AD, 
they mainly mislead the models by manipulating textual inputs, lack physical realizability, and are less robust in the face of real-world changes in lighting, scale, and perspective
The ``Digital'' method in Table~\ref{tab:2}, which leverages PGD combined with feature-level attack, manipulates pixels across the entire image within a norm constraint. They are limited by the perturbation budget and are not suitable for the driving scenarios because VLM-AD predominantly attends to semantic regions such as roads, traffic, and vehicles, which renders a significant portion of perturbed pixels in digital attacks functionally ineffective. 
In contrast, our physical attack focuses on adding adversarial textures directly onto target objects, specifically vehicles in this case.  This attack is more likely to persist under real-world conditions such as varying viewpoints, lighting, and camera noise. 

Our physical attack achieves a significantly higher average ASR (54\%) compared to the Digital method (7\%). This suggests that attacking semantically important, context-aware regions of the vehicle can be more effective in misleading advanced perception and planning modules, especially in complex multi-task driving settings.





\subsection{Ablation Study}

\begin{table}[tbp]
\footnotesize

\caption{Ablation study on Feature Divergence Loss (FDL), sampling strategy, and multi-scale input.}
\vspace{0.3cm}
\centering
\begin{tabular}{l|cccc}
\toprule
\textbf{} & \textbf{PLAINNING} & \textbf{PREDICTION} & \textbf{PERCEPTION} & \textbf{Average} 
\\ \midrule
Encoder FDL              & 61\%              & 36\%               & 17\%                & 38\%           \\
Projector FDL            & 67\%              & 42\%               & 19\%                & 42\%           \\
ML-FDL                   & 69\%              & 44\%               & 23\%                & 45\%           \\
ML-FDL+Sampling          & 73\%              & 51\%               & 27\%                & 50\%           \\
\rowcolor{gray!20}
ML-FDL+Sampling+Multi-scale & 78\%          & 56\%               & 28\%                & \textbf{54\%}          \\ \bottomrule
\end{tabular}
\label{table:3}
\end{table}

\begin{figure}[t!]
    \centering
    \begin{subfigure}[t]{0.54\textwidth}
        \centering
        \includegraphics[width=\linewidth]{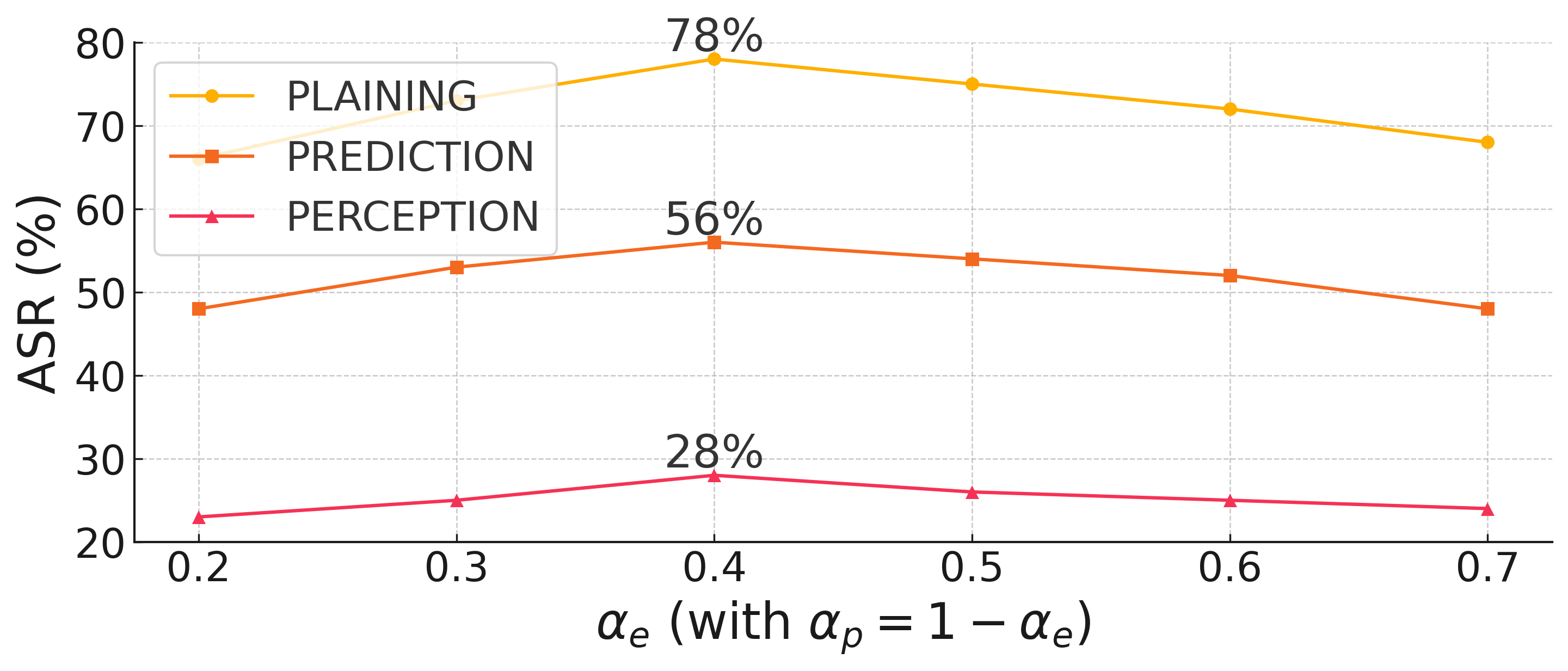}
        \caption{}
        \label{fig:sub5}
    \end{subfigure}
    \begin{subfigure}[t]{0.45\textwidth}
        \centering
        \includegraphics[width=\linewidth]{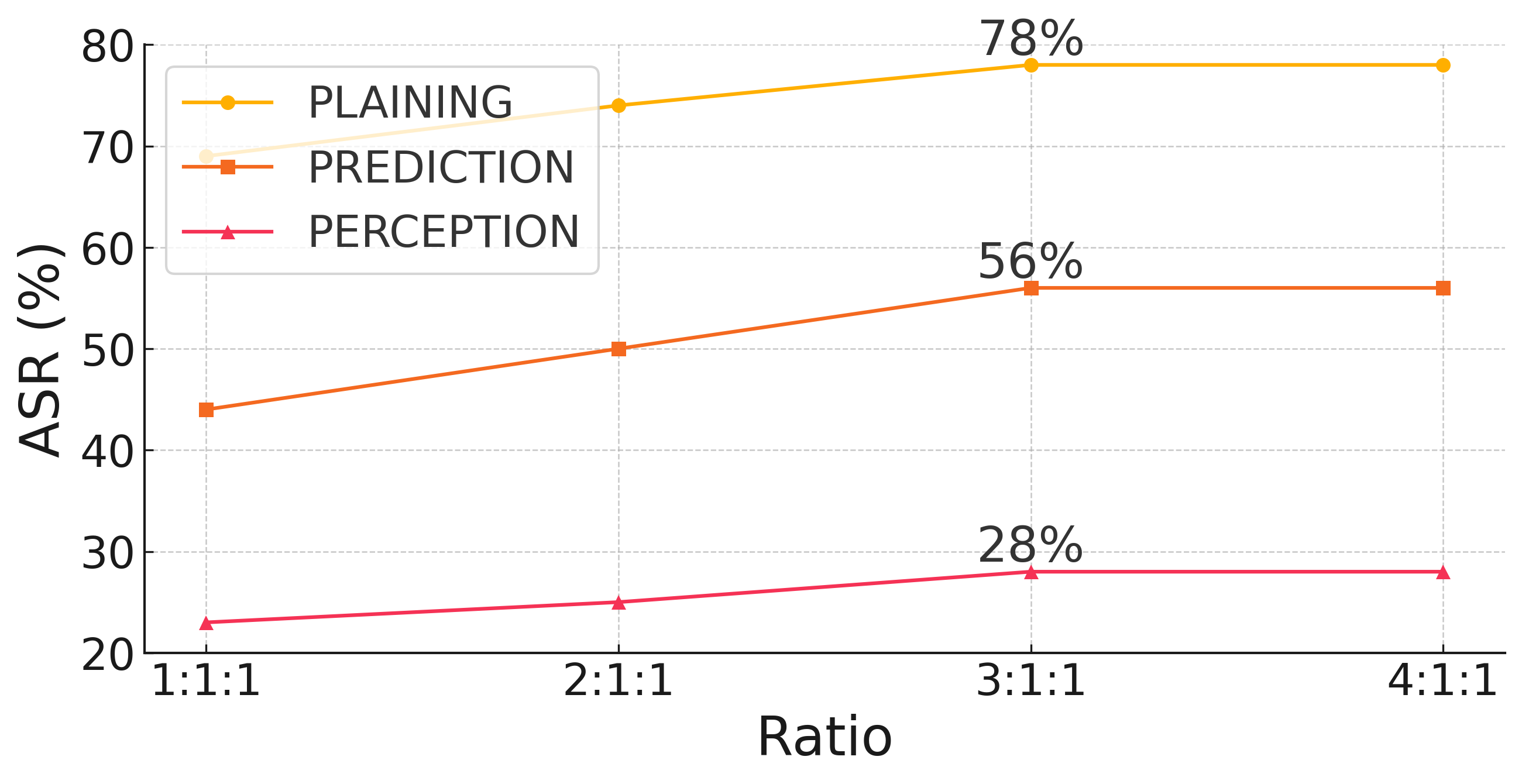}
        \caption{}
        \label{fig:sub6}
    \end{subfigure}

    \caption{Ablation study on weighting factor $\alpha$ and sampling ratio.}
    \label{fig:6}
\end{figure}

we perform ablation studies to analyze the impact of our proposed Feature Divergence Loss (FDL) and physical adaption (Sampling and Multi-scale strategy). Table \ref{table:3} shows the effectiveness of our proposed attack when progressively incorporating various modules. It tells that multi-layer Feature Divergence Loss (FDL) performs better than every single layer loss. Physical adaption is proposed to address the challenge of angle and distance in the real-world environment and improve all driving scenarios, especially Planning and Prediction.

We evaluate the range weighting factor $\alpha$ using a range from 0.2 to 0.7 and sampling ratio by increasing the proportion of 22.5°. Figure \ref{fig:sub5} shows that the best performance was achieved when $\alpha_e=0.4$ and $\alpha_p=0.6$, where  $\alpha_e$ and $\alpha_p$ are weighting factors of the encoder and projector. Figure \ref{fig:sub6} shows that a 3:1:1 sampling ratio achieves better.

%% file: sec/5_conclusion.tex
\section{Conclusion}
\label{sec:con}
Our proposed UCA framework is the first camouflage attack on VLM-AD. By feature divergence and physical adaption, UCA can achieve task-universal attack with strong performance across various distances and angles. We also propose a new 3-P metric and UCA surpasses the state-of-the-art attack by 30\%. However, the attack still primarily succeeds in white-box settings, and improving transferable ability is our future work.